\documentclass[10pt,letterpaper]{article}
\usepackage {amsmath,amsthm,amssymb,latexsym,graphicx}
\usepackage{cogsci}
\usepackage{pslatex}
\usepackage{apacite}
\newtheorem{thm}{Theorem}

\title{A Computational Theory of Subjective Probability}
 
\author{{\large \bf Phil Maguire (pmaguire@cs.nuim.ie)} \\
  Department of Computer Science \\
  NUI Maynooth, Ireland
  \AND {\large \bf Philippe Moser (pmoser@cs.nuim.ie)} \\
  Department of Computer Science \\
  NUI Maynooth, Ireland
\AND {\large \bf Rebecca Maguire (rebecca.maguire@ncirl.ie)} \\
  School of Business, National College of Ireland \\
  Mayor Street, IFSC, Dublin 1, Ireland
\AND {\large \bf Mark Keane (mark.keane@ucd.ie)} \\
  School of Computer Science and Informatics \\
  UCD, Dublin 4, Ireland}

\begin{document}

\maketitle

\begin{abstract}
In this article we demonstrate how algorithmic probability theory is applied to situations that involve uncertainty. When people are unsure of their model of reality, then the outcome they observe will cause them to update their beliefs. We argue that classical probability cannot be applied in such cases, and that subjective probability must instead be used. In Experiment 1 we show that, when judging the probability of lottery number sequences, people apply subjective rather than classical probability. In Experiment 2 we examine the conjunction fallacy and demonstrate that the materials used by \citeA{TverskyKahneman1983} involve model uncertainty. We then provide a formal mathematical proof that, for every uncertain model, there exists a conjunction of outcomes which is more subjectively probable than either of its constituents in isolation.  

\end{abstract}

\section{Introduction}

\emph{Breaking news: Pandemonium erupted today at the National Lottery headquarters as the numbers 1, 2, 3, 4, 5 and 6 were drawn for the third week in a row. Lottery officials, stunned by a sense of d\'{e}j\`{a} vu, scrambled to release a statement insisting that the lottery drum selection mechanism meets the highest standards for randomness. Meanwhile thousands are celebrating after ignoring the opinions of mathematicians who had viewed the two previous draws as a statistical fluke. Commentators in the media are demanding an immediate investigation, describing the incident as a fiasco.}

The mathematical concept of probability, originally formulated to describe the highly constrained environment of games of chance, has now found its way into everyday parlance, with people using it to quantify the likelihood of everything from the possibility of economic recession to the risk of global warming. Such has been the unquestioned adoption of the probability concept into mainstream culture that it has become the default assumption that probability theory provides the only logical way for people to think about likelihood.
For instance, \citeA{TverskyKahneman1983} applied probability theory to real-world situations involving personality decisions, medical judgments, criminal motives and political forecasts. On observing consistent deviations from the mathematical theory, they interpreted their findings as evidence of a serious flaw in human reasoning (see \citeNP{Costello2008}, for a review of the associated debate). In this article we adopt the alternative stance that consistent deviations between human reasoning and a simplified, artificial mathematical theory are far more likely to reflect deficiencies in the theory than they are to reflect sub-optimality in how people think about likelihood.

\section{Classical Probability}

Probability theory was formalised by Kolmogorov in the 1930s through the notion of probability space, whereby a set of possible outcomes is mapped to a number that represents its likelihood by a probability measure function. For example, a perfect dice outputs the numbers from 1 to 6 with equal frequency. 
However, in the real world it is rarely feasible to identify the theoretical probability measure function which underlies the events we observe. Because we have to work backwards, using the events to deduce the original function, we can never be sure if the model we are using is correct.

For example, according to classical probability theory, no conceivable sequence of numbers produced by rolling a dice will ever lead us to revise our beliefs about the nature of the dice. Even if we rolled 1, 1, 1, 1... the hypothesis of the sequence being a statistical fluke would always remain infinitely more likely than the possibility that the dice is biased.

In reality, nobody has beliefs which are strong enough to stand up to the requirements of classical probability theory. We strongly believe that the numbers drawn from the lottery are random, yet there are certain sequences which, as in the introductory lottery example, would cause us to question our assumptions and consider other possibilities. If the sequence 1, 2, 3, 4, 5, 6 was drawn three weeks in succession, it might suggest that the balls were not equally weighted, the drum mechanism was defective, or that one of the lottery officials was playing a practical joke. The point where we start to ask questions reveals how strongly we hold our beliefs. But no matter how confident we are about a particular model of reality, there will always be some sequence of events which will cause us to change our mind.

This poses a crucial problem for probability theory. Let's consider the probability of 1, 2, 3, 4, 5, 6 being drawn in a lottery for three weeks in a row. If the draw is unbiased then this sequence of events is just as likely as any other. In a lottery with 45 numbers, the exact probability is $C(45,6)^3$. But if this sequence of events actually unfolded, it would lead us to believe that the draw mechanism is biased. Given the new updated belief, then the probability of getting 1, 2, 3, 4, 5, 6 is actually far higher. So what is the true probability of this sequence of events?

To apply classical probability theory a single model of reality must be selected. We must assume either that the lottery draw is biased or that it isn't. But doing so would be a mistake because we don't actually know which world is the case. The situation involves model-outcome dependence, insofar as the outcome affects our beliefs about the system that generated it. Stating that the probability of drawing 1, 2, 3, 4, 5, 6 is $C(45,6)^3$ is misleading because, if this sequence of events actually occurred, we would no longer trust the assumptions involved in computing that probability.

\section{Uncertainty in the Real World}

The issue here is that classical probability theory only applies to cases involving a definitive probability measure function, while models of reality always involve uncertainty. Though useful for reasoning about games deliberately engineered to generate pseudo-randomness, classical probability has less applicability to everyday life, where reducing uncertainty and optimising models of reality are the principal goals. 
In our previous work examining the difference between surprise and probability judgments \cite{ Maguire2011} we presented a cognitive theory of uncertainty modeling which views the maintenance of an up-to-date representation of reality as the principal motivation guiding information seeking behaviour. People rely on observational data to continually refine their model of the environment, thus maintaining the optimality of their decision making. In particular, the signal that they rely on to diagnose discrepancies between their model and the real world is randomness deficiency. 

The best model of a set of observational data is the one which describes it most concisely, so that the description of the data relative to the model is `incompressible' or random (see \citeNP{Rissanen1978}; \citeNP{Gacs2001}). In the case that one's model of reality is optimal, then new sensory data should still be random with respect to it. The experience of randomness deficiency (i.e. a pattern which could be described more concisely using an alternative model) causes alarm bells to go off, because it indicates that one's model is likely to be suboptimal. This is known as surprise.

When surprise occurs there are two potential resolutions. First, more observation data can be gathered, which might mitigate the randomness deficiency by revealing it to be a statistical fluke. If this does not resolve the discrepancy then the remaining alternative is to update one's model to fit the data. Either way, the resolution process necessitates urgent sampling of information from the environment. During the surprise response, eye widening, opening of the mouth and enlargement of the nasal cavity serve to facilitate the intake of sensory information (see \citeNP{Maguire2011}). 

Consider for example looking at the floor and seeing some crumbs which spell out the words ``YOU ARE BEING WATCHED''. When crumbs fall on the floor it is just as probable that they will arrange themselves into this pattern as any other. If we were certain that the crumbs had fallen randomly then it would not be interesting. However, where knowledge is uncertain then people respond to randomness deficiency. The pattern of crumbs is randomness deficient because there is another model which can explain it more concisely: Somebody might have deliberately arranged the crumbs in this way. The first strategy is to look at the rest of the floor. If the rest of the floor is covered in many crumbs which have no other patterns then the overall randomness deficiency is mitigated. If these are the only crumbs on the floor then finding a satisfactory explanation becomes critical. 

People are motivated to seek out randomness deficiency in the world \cite{Dessalles2006}. The experience of randomness deficiency with subsequent resolution through representational updating is what makes subjects interesting, films entertaining and jokes funny \cite{Schmidhuber2009}. Accordingly, when people speak intuitively about likelihood and probability, it is the concept of representational updating which is relevant to them.

\section{Subjective Probability}

Because it assumes a definitive probability measure function, classical probability theory cannot be applied to the concept of representational updating. This limitation means that the theory is, for the large part, irrelevant to everyday life and thus inappropriate for evaluating the nature of human reasoning. 

Developments in algorithmic statistics have allowed probability theory to be extended to situations involving an uncertain probability measure function (see, e.g., \citeNP{Vityanyi2000}; \citeNP{Gacs2001}). The optimal model which can be derived from a set of observations is the one which maximizes the compression of that dataset, yielding the Minimum Description Length (MDL), a concept which formalizes Occam's razor. 

Whenever an observation is no longer typical with respect to an MDL model it should be adjusted to lower the randomness deficiency of the data (see \citeNP{Li2008}, for details on how the updating process is carried out). We can quantify the extent of this representational adjustment in terms of the amount of information that, given the original model, would be required to obtain the updated model. The more the information required, the more significant (and less likely) the update.

The model that people hold of reality represents the very best that they can do in representing their environment and provides the very best that they can achieve in terms of predictions. If we assume that our representation is a reliable predictor of events then the larger a potential update to that representation, the rarer it should be. Accordingly, we can apply probability theory to speak about the likelihood of an outcome requiring an update of a particular size. The uncertainty which precludes probability theory from being applied to real-world scenarios is circumvented by shifting the focus from an underdetermined probability measure function to the immutable mechanism of representational updating.

\subsection{Preliminaries}

A computable probability density function $p$ can be interpreted as a model for a string generating device. Given such a device, described by $p$,
there are some ``type of strings'' we expect to be output, whereas some others are surprising. 
String $x$ is said $p$-typical if it is a random string relative to the model described by $p$,
i.e. the model already describes all the regularities in $x$.

Formally, let $\alpha >0$ be a constant, called the surprise threshold, which represents the level of randomness deficiency that necessitates representational updating.
String $x$ is $p$-typical with surprise threshold $\alpha$ (or $(p,\alpha)$-typical) if the length of its shortest description given $p$ is 
at least the number of bits a Shannon-Fano code based on $p$ would require (an encoding where the more 
$p$-likely a string is, the shorter its encoding will be) after subtracting the surprise level $\alpha$, i.e.,

$$K(x|p^*) \geq -\log p(x) - \alpha.$$

The idea behind the minimal description length (MDL) of a string $x$ \cite {Gacs2001} is to take the shortest (in description length)
among all models for which $x$ is typical.
To avoid overfitting (i.e. the model is specifically built for $x$ instead of for all ``strings of type $x$'') 
the description length of both the model and the string given the model,
should be equal to the description of the string on its own.
Formally, probability density function $p$ is optimal for string $x$ if the shortest description of $x$ has the same length (up to an additive constant)
as the shortest description of $p$ plus the number of bits required for a Shannon-Fano encoding of $x$ based on $p$, i.e.,
$$K(x) = K(p) -\log p(x) \pm O(1)$$
where $O(1)$ means the equality holds up to an additive constant.
The MDL of string $x$ is the shortest (description length) among all optimal probability density functions for $x$ for which $x$ is typical.

\subsection{Subjective information and probability}

Suppose an observer experiences observations
$d_1,d_2,\ldots$ generated by some source with computable probability density $p_{\mathrm{source}}$.
The observer  tries to learn the probability density $p_{\mathrm{source}}$ by finding the shortest optimal model based on the observations made so far.
Formally, after having observed strings $d_1,d_2,\ldots,d_n$, the observer seeks to construct a hypothetical model $p_n$ where

\begin{align*}
p_n &= \arg\min\{|p^*|: p \text{ is optimal for } 
d_1,d_2,\ldots,d_n \text{ and }\\
&d_1,d_2,\ldots,d_n \text{ are } (p,\alpha)\text{-typical}\}.
\end{align*}

If the next observation $d_{n+1}$ is surprising, action may be required.
Formally, observation $d_{n+1}$ is $\alpha$-surprising if  the length of its shortest description given $p$ is 
less than the number of bits a Shannon-Fano code based on $p$ would require 
after subtracting the surprise level $\alpha$, i.e.,
$$K(d_{n+1}|p_n^*)<-\log p_n(d_{n+1}) - \alpha.$$

If an update is performed, then the subjective information of $d_{n+1}$ (the ``cost'' of the update) is 
the amount of information needed to update the model to the latest, that is the length of the shortest 
description of the new model, given the old model, i.e., 
$$\mathrm{subjective \ information}(d_{n+1}) = K(p_{n+1}^*|p_n^*).$$

Subjective probability (the probability of the update) can then be quantified based on the amount of information it contains, i.e.,

$$\mathrm{subjective\ probability}(d_{n+1}) =2^{-K(p_{n+1}^*|p_n^*)}.$$

\section{Experiment 1}

In the following experiment we investigated the hypothesis that people use subjective probability rather than classical probability to judge the likelihood for real-world events. We used an example for which the use of classical probability theory seems particularly compelling, namely lottery sequences (see \citeNP{Dessalles2006}). A naive application of classical probability suggests that all lottery sequences are just as likely. 

\subsection{Method}

In a lottery system where 6 numbers are drawn from 45, each ordered sequence has a classical probability of $C(45,6)$. According to the theory outlined in the previous section, the subjective probability of an outcome is related to its randomness deficiency. People expect the lottery numbers to be Kolmogorov-random. The more they deviate from a typical random string, the lower the subjective probability that they reflect the output of a random source. The randomness deficiency of a string is quantified precisely by its MDL. However, since this theoretical construct is not computable in practice, we are obliged to create a heuristic compressor which approximates it. 

We considered the patterns to which people are sensitive in discriminating predictable sequences from random ones. Overtly non-typical random patterns include ones in which the numbers are consecutive (e.g. 3, 4, 5, 6, 7, 8) or where they increase in a constant step size. To compress these patterns we created a simple compressor which takes in an ordered sequence of six numbers, and computes the six step sizes between them (with the first number counting as the first step). A Huffman encoding scheme is then applied, which relates bit size to step size. A breakdown of the structure of the associated Huffman tree is provided in Table~\ref{huffman}.

Using this system the sequence 10, 32, 33, 35, 39, 45 is transformed to step sizes of +10, +22, +1, +2, +4, +6 which is then encoded using 8 + 8 + 2 + 3 + 4 + 6 = 31 bits. Analysing six years of bi-weekly Irish National Lottery draws revealed a mean compressed length of 30.9 bits, with a mode of 31 bits. The most randomness deficient of the 624 sequences was 2, 4, 32, 34, 36, 37 (description length of 20 bits), while the most random was 9, 20, 26, 27, 34, 45 (description length of 39 bits). The theoretical minimum description length of our system was 12 (e.g. 1, 2, 3, 4, 5, 6), while the theoretical maximum was 43 (e.g. 7, 13, 20, 29, 36, 45). The number of bits needed to perfectly encode an ordered random sequence of six numbers between 1 and 45 is 23.0 bits. Although our compressor cannot compute MDL, it delivers compression for randomness deficient outputs (i.e. it compresses below 23.0 bits for certain non-typical random sequences) and can therefore be used to evaluate the hypothesis that people use subjective rather than classical probability.

\begin{table}[!ht]
\begin{center} 
\caption{Structure of Huffman encoding scheme.} 
\label{huffman} 
\vskip 0.12in
\begin{tabular}{lll} 
\hline
Level Depth    &  Leaves & \#Branches \\
\hline
1        &   - & 2  \\
2   &   +1, repeat & 2 \\
3           &   +2, +3 & 2 \\
4           &   +4 & 3 \\
5           &   +5 & 5 \\
6           &   +6 & 9 \\
7           &   +7, +8 & 16 \\
8           &   +9 up to +40~~~~ & - \\
\hline
\end{tabular} 
\end{center} 
\end{table}

\subsubsection{Participants}

130 undergraduate students from NUI Maynooth participated voluntarily in this study.

\subsubsection{Procedure}
As an initial step we purchased two quickpick (i.e. randomly selected) lottery tickets for the next week's Irish National Lottery, with six ordered numbers ranging from 1 to 45. 
Participants were informed that we had purchased these tickets and that, for each of the two quickpick sequences, their goal was to identify it from among a group of five candidate sequences. No mention was made of how the other four sequences had been generated.

Each quickpick sequence was presented on a screen along with four other sequences randomly generated using our compressor algorithm. The four distractor sequences met the constraints of having compressed bit-sizes of between 15 and 18 bits, 19 and 22 bits, 23 and 26 bits, and 27 and 29 bits respectively. As it happened, the first lottery ticket sequence had a compressed description length of 31 bits, and the second had a length of 30 bits. The ordering of the five sequences on the screen was randomized.

Participants ranked each set of five sequences in order of likelihood of being the quickpick sequence, from highest probability to lowest probability. After the process was complete participants were shown the actual lottery tickets so that, as promised, they could see if they had made the correct judgment or not.

Unfortunately for the experimenters, the lottery tickets did not turn out to be winning ones.

\subsection{Results and Discussion}
An individual applying classical probability would view all sequences as equally likely and would thus only have a 20\% chance of correctly identifying one quickpick sequence mixed with four others. However, 64\% of participants correctly identified the numbers on the first ticket, and 66\% on the second ticket (i.e. ranked these sequences in first place). When participants were shown the lottery tickets at the end of the experiment they were surprised that their intuition had, in the majority of cases, led them to make the correct choice.

\begin{figure}[ht]

\begin{center}

\includegraphics[width=8cm]{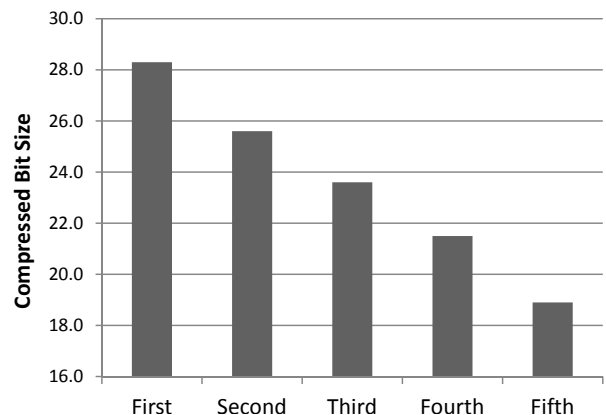}

\end{center}
\caption{Mean compressed bit size according to rankings of likelihood.} 
\label{experiment1results}
\end{figure}

Figure~\ref{experiment1results} shows the mean compressed bit size for sequences ranked from first to fifth place across the two presentations. The overall correlation between ranking and compressed description length was 0.965, $p < .001$.

These results demonstrate that, not only do people use subjective probability, they also enhance the accuracy of their judgments by using it. While the naive mathematician assumes all lottery sequences are equally likely, the savvy layperson realises there is an element of uncertainty involved in how those sequences were generated. The greater the randomness deficiency of a sequence, the greater the subjective probability that it was generated by a non-random generative mechanism. 

Our central argument in this article is that, because models of reality always involve uncertainty, people apply subjective probability rather than classical probability in everyday life. In the following experiment we investigated whether the application of subjective probability can explain experimental observations which have previously been interpreted as examples of fallacious reasoning. 

\section{Experiment 2}

The conjunction effect is a situation in which people assert that a conjunction of two outcomes is more probable than either of those outcomes in isolation. According to classical probability theory this is a fallacy because requiring two outcomes to be validated is always a stricter criterion than requiring a single one to be validated (i.e. $P(x\wedge y) \leq P(y)$). The most celebrated example of the fallacy involves one of the materials used by \citeA{TverskyKahneman1983}, involving an individual named Linda.

\emph{Linda is 31 years old, single, outspoken, and very bright. She majored in philosophy. As a student, she was deeply concerned with issues of discrimination and social justice, and also participated in anti-nuclear demonstrations}.

Which is more probable?

a) Linda is a bank teller

b) Linda is a bank teller and is active in the feminist movement.

\citeA{TverskyKahneman1983} report that, when the two possible outcomes are listed together as above, 85\% of people violate the conjunction rule by identifying b) as more probable. \citeauthor{TverskyKahneman1983}'s explanation is that people get confused by what they call `representativeness'. They found that participants' responses reflect the extent to which the descriptions match a stereotype, with a correlation of 0.98 between mean ranks of probability and representativeness. 

It is interesting to note that this correlation closely matches the observed correlation of 0.97 between mean ranks of probability and compressed description length in Experiment 1. This suggests the possibility that representativeness and randomness deficiency are closely related concepts. 

In Experiment 1 we found that, when there is uncertainty as to the generative mechanism which produced an outcome, people rely on randomness deficiency to make judgments. The uncertainty in Experiment 1 concerned the fact that participants were given no information as to how four of the five lottery sequences were generated. Rather than assuming that all the sequences were generated randomly, they correctly used randomness deficiency to make inferences that resolved the uncertainty.

In the Linda example, some information about Linda is provided, but there is much about her that remains unknown (e.g. has she settled down since her student days?) In the case of uncertainty regarding the underlying probability measure function, then classical probability cannot be applied. For example, if we find out that Linda is a bank teller, then we might infer that she has settled down. In contrast, hearing that she is still active in the feminist movement suggests that she has not changed much since her student days. Because these two models of Linda are quite different, there is no definitive probability measure function relative to which classical probability can be expressed.

\subsection{Method}

In the following experiment we investigated whether the outcomes for the Linda scenario cause participants to adjust their model of Linda.  

\subsubsection{Materials}

For this experiment we altered the Linda scenario by including the outcomes as part of the description. We removed the information that she is single, outspoken and very bright and included at the end of the description either that ``Linda is a bank teller'' (Version 1) or ``Linda is a bank teller and is active in the feminist movement'' (Version 2). Participants were then asked to rate the probability of Linda having the attributes of being single, outspoken and very bright (from 0 to 100\%). In order for classical probability to be applicable, then the probabilities provided for Versions 1 and 2 should not differ significantly. Linda should be just as independent, outspoken and bright regardless of whether she is active in the feminist movement or not.  

\subsubsection{Participants}

106 undergraduate students from NUI Maynooth participated voluntarily in this study.

\subsubsection{Procedure}

Participants were randomly assigned either Version 1 or Version 2 of Linda's description and wrote down their probabilities for the three characteristics, which were randomly ordered along with three other filler characteristics (Linda plays golf, Linda is dyslexic, Linda suffers from anxiety).

\subsection{Results and Discussion}

The mean probabilities for the three characteristics are shown in Table~\ref{lindaresults}. When Linda was described as a bank teller and active in the feminist movement she was rated as significantly more likely to be single, demonstrating that the outcomes used in the Linda scenario cause participants to adjust their model of Linda. 

\begin{table}[!ht]
\begin{center} 
\caption{Mean probability ratings, t-test scores and significance for the two descriptions of Linda.} 
\label{lindaresults} 
\vskip 0.12in
\begin{tabular}{llll} 
\hline
    &  Ver. 1 & Ver. 2 & t-test \\
\hline
Single        &   47\% & 64\% & $t(104) = 4.11, p <.001$ \\
Outspoken   &   77\% & 80\% & $p > .05$ \\
Very Bright           & 59\% & 63\% & $p > .05$    \\
\hline
\end{tabular} 
\end{center} 
\end{table}

The numbers generated by a perfect dice never lead us to update our beliefs about the nature of the dice, yet finding out about Linda's current activities does lead people to update their beliefs about her. Because the model of Linda is uncertain, subjective probability must be applied. What people are quantifying when they identify the conjunction as more probable is that the conjunction contains more subjective information, and that, relative to the process of representational updating, the likelihood of an outcome diminishes with the amount of subjective information it carries. Basing decisions on subjective probability is mathematically the correct approach when dealing with uncertainty regarding the underlying probability measure function.

In the following section we build on this result by proving that for every situation involving uncertainty (i.e. all real world scenarios) there is a conjunction of events which is more subjectively probable than either of its constituents in isolation. 

\section{Proof that Conjunction Effect is not a Fallacy}
In this section we prove that given any hypothetical model $p$, there are always two strings of events $x,y$
such that $x$ is a substring of $y$ but $y$ has higher subjective probability. The idea of the proof is that any long enough typical string of events can always be decomposed into 
a substring of events that carries greater subjective information.

\begin{thm}
Let $E_1,E_2,\ldots E_m$ be $m$ independent events and let $p$ be the associated computable probability measure function. 
Let $\alpha>0$ be a surprise threshold. There exists a conjunction of events $A=A_1 \wedge A_2 \wedge\ldots\wedge A_n$ with a constituent $B$
(i.e. $p(A)<p(B)$) such that
$B$ is $(p,\alpha)$-surprising (i.e. carries subjective information) and $A$ is $(p,\alpha)$-typical (i.e. has a subjective probability of 1).

\begin{proof}
Let $E_1,E_2,\ldots E_m$,  $p$ and
$\alpha>0$ be as above. Without loss of generality $m=2^k$ and $p$ can be seen as a probability on strings of length $k$ (each coding one event $E_i$) extended multiplicatively
i.e.,
$p:2^{k}\rightarrow [0,1]$ is extended multiplicatively by $p(xy):=p(x)p(y)$. 

Let $n$ be a large integer. Let $y\in 2^{kn}$ be a $(p,\alpha)$-typical string.
$y$ can be viewed as the concatenation of $n$ strings of length $k$ (i.e. the conjunction of $n$ events). 
By the pigeon hole principle,
there must be such a string that occurs at least $n/2^k$ times. Denote this string by $s$, and let $l$
be the number of occurences of $s$ in $y$, i.e. $l\geq n/2^k$.
Because $y$ is $(p,\alpha)$-typical we have $p(s)>0$.
Thus $p(s)=2^{-c}$ for some $c>0$. 
Let $x$ be $l$ concatenations of $s$. Because $p$ is extended multiplicatively we have $p(x)>p(y)$.

Let us show that $x$ is $(p,\alpha)$-surprising.
To describe $x$ it suffices to describe $l$ plus a few extra bits that say ``print $s$ $l$ times''.
Since $l$ can be described in less than $2\log l$ bits (by a prefix free program) we have
$K(x) < 3 \log l$ for $n$ large enough.
We have

\begin{align*}
-\log p(x) - \alpha &= -\log p(s^l) - \alpha = -\log p(s)^l -\alpha\\ 
&= -l\log 2^{-c} - \alpha 
= cl -\alpha > 3 \log l > K(x) \\
&\geq K(x|p^*)
\end{align*}

for $n$ large enough. Thus $x$ is $(p,\alpha)$-surprising, but $y$ is not.
\end{proof}
\end{thm}

\section{Conclusion}

Although \citeA{TverskyKahneman1983} identified an association between representativeness and the conjunction effect, they never provided an explanation for why such an association might exist, instead being satisfied to pass it off as an arbitrary reasoning fallacy. Had they questioned participants regarding their judgments, rather than dismissing them as fallacious, then the resultant findings may have facilitated the extension of classical probability theory. In sum, perhaps the most salient fallacy on display in \citeauthor{TverskyKahneman1983}'s \citeyear{TverskyKahneman1983} study is the misplaced belief that mathematical theories which have been developed for precision models in the exact sciences retain their validity when used to describe complex cognition in the real world. 

\citeA{TverskyKahneman1983} posed the following question: ``Why do intelligent and reasonably well-educated people fail to recognize the applicability of the conjunction rule in transparent problems?'' Here, we have presented the answer: Because often it's not applicable.

\bibliographystyle{apacite}

\setlength{\bibleftmargin}{.125in}
\setlength{\bibindent}{-\bibleftmargin}

\bibliography{References}

\end{document}